\documentclass[letter,twocolumn]{jpsj3}
\usepackage{txfonts,algorithm, algorithmic,color}

\title{Belief Propagation for Maximum Coverage on Weighted Bipartite Graph and Application to Text Summarization}

\author{Hiroki Kitano and Koujin Takeda$^1$\thanks{koujin.takeda.kt@vc.ibaraki.ac.jp}}
\inst{$^1$Department of Mechanical Systems Engineering, Graduate School of Science and Engineering, Ibaraki University\\
4-12-1 Nakanarusawa, Hitachi, 316-8511 Ibaraki, Japan \\} 

\abst{We study text summarization from the viewpoint of maximum coverage problem.
In graph theory, the task of text summarization is regarded as maximum coverage problem on bipartite graph with weighted nodes.
In recent study, belief-propagation based algorithm for maximum coverage on unweighted graph was proposed
using the idea of statistical mechanics.
We generalize it to weighted graph for text summarization. Then we
apply our algorithm to weighted biregular random graph for verification of maximum coverage performance. 
We also apply it to bipartite graph representing real document in open text dataset, and
check the performance of text summarization.
As a result, our algorithm exhibits better performance than greedy-type algorithm in some setting of
text summarization.}

\begin{document}
\maketitle

Text summarization (TS) is one of the important tasks in natural language processing, and
many TS methods have been proposed. 
Among them, we focus on the summarization method 
to exclude as many redundant sentences in the document as possible.
In such method,
TS is regarded as an optimization problem.
For example, TS is reformulated as knapsack problem in the past study\cite{McDoland}, 
where global optimal solution or approximation solution is discussed. 
TS can also be viewed as maximum coverage (MC) problem of nodes in graph theory,
as first discussed in Ref.\citen{Filatova}. In their work, simple greedy algorithm is used to find approximate solution of MC, 
because MC is NP-hard. Hence, there may exist more appropriate algorithm for MC than simple greedy algorithm. Actually,
several MC algorithms are compared in the previous work.\cite{TO}

In statistical mechanics, optimization such as MC is regarded as the problem to find ground state of system.
In Ref.\citen{TMH}, they proposed a novel MC algorithm based on belief propagation (BP), 
where additional physical parameters, i.e. temperature and chemical potential, are introduced to control optimization.
As a result, they could find better solution than greedy algorithm by tuning physical parameters.
However, their algorithm is for unweighted bipartite graph. In order to apply this algorithm to TS, 
generalization to weighted graph is necessary.

From such background, we consider MC on weighted bipartite graph
for TS.
First we give BP-based MC algorithm for weighted bipartite graph.
Then we conduct MC experiment on biregular random graph in order to compare with an improved greedy algorithm
for weighted graph\cite{KMN}.
Next we apply our algorithm to real document data\cite{DUC4},
and evaluate the performance of TS quantitatively.


Here we formulate MC on weighted bipartite graph.
We separate the nodes into two groups ${\cal X},{\cal Y}$ on bipartite graph, where the nodes
in different groups are not directly connected. 
The numbers of elements are $|{\cal X}|=N$ and $|{\cal Y}|=M$ respectively, where $| \cdot |$ means cardinality.
The set of edges between ${\cal X}, {\cal Y}$ is denoted by $\cal E$.
The binary variable $x_i \in \{ 0,1 \}$ is defined on the $i$th node in $\cal X$, and
$y_{a} \in \{0, 1\}$ on the $a$th node in $\cal Y$. We also define weight for each nodes,
$c_i$ in $\cal X$ and $w_{a}$ in $\cal Y$. Our objective is to solve
the integer programming for MC as

\begin{figure}[t]
\begin{picture}(0,158)
\put(15,-5){\includegraphics[width=0.40\textwidth]{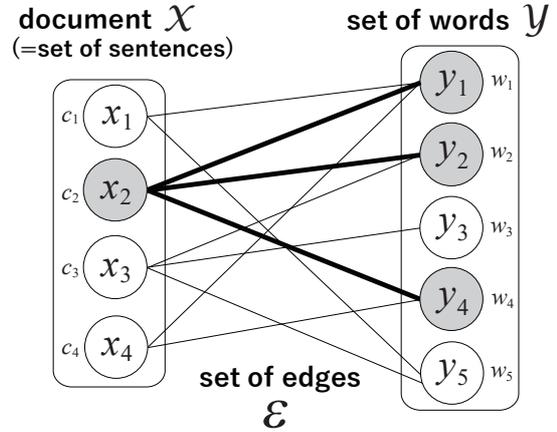}}
\end{picture}
\caption{MC on weighted bipartite graph and TS: The left node $x_i \in {\cal X}$ is a sentence, while
the right node $y_a \in {\cal Y}$ is a word. The weight of the left node $ c_i $ is the number
of words in the sentence, while the one of the right node $w_a$ describes importance of a word. 
Sentences are selected to cover as much weight in connected nodes (=words) as possible (shaded nodes in
the figure)
with the upper bound for the number of words.}
\label{f1}
\end{figure}

\begin{align}
& {\rm maximize} \ \ \sum_{a} w_{a} y_{a} \nonumber \\
& s.t. \sum_i c_i x_i \le K, \ \ y_{a} \le \sum_{i \in \partial a} x_i\ (\forall a),
\label{eq:opt}
\end{align}
where $K$ is parameter for upper bound of constraint.
The last inequality for $y_a$ means that $y_{a}=1$ if $x_i = 1 \ \exists i \in \partial a$, where
$\partial$ represents neighborhood.
The value $x_i=1$ means the $i$th node is selected for covering connected nodes in $\cal Y$, while $y_{a}=1$ represents at least
one of $a$'s connected nodes is selected for coverage. If $c_i = w_{a} = 1 \ \forall i, a$, 
this integer programming is reduced to unweighted MC. In this case, $K$ nodes
 in $\cal X$ are selected to cover as many connected nodes in $\cal Y$ as possible, and the performance of coverage is measured by
$\sum_{a} y_{a}$. See also Fig.\ref{f1}. 

In the context of TS,
each node in $\cal X$ is taken as a
sentence in the document, and each node in $\cal Y$ corresponds to a word.
The weight $c_i$ means how many words the $i$th sentence includes,
and the weight $w_\mu$ describes the importance of the $\mu$th word.
Using the integer programming in Eq.(\ref{eq:opt}), we want to
cover as much weight of words as possible by selecting significant sentences 
in the document, under the condition that the number of total words in the selected
sentences is smaller than $K$.


The problem in the current TS framework is that the integer programming in Eq.(\ref{eq:opt}) is NP-hard.
Hence we need an algorithm for good approximate solution.
In Ref.\citen{KMN}, the integer programming is solved approximately by greedy algorithm with performance guarantee
in Algorithm \ref{alg1}, called g-greedy hereafter.
This was applied to TS, and found to show good performance in comparison with
other algorithms\cite{TO}. 
In this algorithm, we select the additional node $i$ in ${\cal X} = \{ 1, 2, \ldots, N \}$ to
maximize the weight sum of connected ($\in \partial i$) and uncovered ($\in \partial ( {\cal X} - \hat{\cal X}_{\rm cov})$) 
nodes in ${\cal Y} = \{ 1, 2, \ldots, M \}$
divided by its weight $c_i$, i.e. $\sum_{ a \in \partial i \cap \partial ( {\cal X} - \hat{\cal X}_{\rm cov})} w_a / c_i$.
In contrast, the algorithm without the weight $c_i$ in the third line in Algorithm \ref{alg1}, i.e. 
$k = {\rm argmax}_{i \in \hat{\cal X}} \{ (\sum_{ a \in \partial i \cap \partial ( {\cal X} - \hat{\cal X}_{\rm cov})} w_a ) \}$,
is called (simple) greedy algorithm in this letter.

\begin{algorithm}                      
\caption{g-greedy algorithm}         
\label{alg1}                    
\begin{algorithmic}                  
\STATE{initialize two sets, $\hat{\cal X} = \{1,\ldots N\},\hat{\cal X}_{\rm cov}= \phi$}
\WHILE{$\hat{\cal X} \ne \phi$}
\STATE{$k = {\rm argmax}_{i \in \hat{\cal X}} \{ (\sum_{ a \in \partial i \cap \partial ( {\cal X} - \hat{\cal X}_{\rm cov})} w_a ) / c_i \}$}
\IF{$c_k + \sum_{i \in \hat{\cal X}_{\rm cov} } c_i \le K$}
\STATE{add $k$ to $\hat{\cal X}_{\rm cov}$}
\ENDIF
\STATE{delete $k$ from $\hat{\cal X}$}
\ENDWHILE
\STATE{output $\hat{\cal X}_{\rm cov}$ (=selected nodes in $\cal X$)}
\STATE{output $\sum_{a \in \partial \hat{\cal X}_{\rm cov}} w_a$ (=weight sum of covered nodes in $\cal Y$)}
\end{algorithmic}
\end{algorithm}

For better solution of Eq.(\ref{eq:opt}) than g-greedy algorithm, we construct BP algorithm. 
The original idea to apply BP to MC is 
proposed in Ref.\citen{TMH}, where weight on the graph is not taken into 
consideration. Hence
we must generalize BP to weighted model in order to apply their idea to the current problem.

Following Ref.\citen{TMH}, we define the partition function for
MC on weighted bipartite graph from Eq.(\ref{eq:opt}),
\begin{align}
Z(\beta) & = \sum_{x_1,\ldots x_N} \sum_{y_1, \ldots, y_M}
\exp \left\{ \beta \left( \sum_{a=1}^M w_a y_a - \mu \sum_{i=1}^N c_i x_i \right) \right\} \nonumber \\
& \hspace{3cm} \times \prod_{a=1}^M \theta \left( \sum_{i \in \partial a} x_i - y_a \right),
\label{eq:partition}
\end{align}
where $\beta$ is inverse temperature, $\mu$ is chemical potential, and $\theta$ is Heaviside function.
As commented in Ref.\citen{TMH}, the constraint $\sum_i c_i x_i \le K$ is not directly incorporated
because it will make the algorithm infeasible. 
Instead, $\mu$ is introduced as an additional control parameter, which also serves as
Lagrange multiplier.
In the limit of $\mu \to \infty$, g-greedy algorithm is reproduced.
Another parameter $\beta$ serves as the relaxation parameter of optimization.

From this partition function, we want to calculate the marginal probabilities,
\begin{eqnarray}
P_i (x_i) \propto \exp \left( \beta h_i x_i \right),\ P_a (y_a) \propto \exp \left( \beta \eta_a y_a \right)
\label{eq:prob}
\end{eqnarray}
to know which node in $\cal X$ should be selected for MC.
The variables $h_i, \eta_a$ are local fields in physical meaning, and
BP is used to calculate these fields.
The generalization of algorithm in Ref.\citen{TMH} to our weighted case is straightforward, 
and the final update algorithm of beliefs is obtained as 

\begin{eqnarray}
h_{ia} &=& - \mu c_i + \sum_{b \in \partial i \backslash a} \hat{h}_{bi}, \label{eq:BP1} \\
\hat{h}_{ai} &=& -\frac{1}{\beta} \ln \left\{ 1 - \frac{1}{1+e^{-\beta w_a}} \frac{1}{\prod_{j \in \partial a \backslash i}
(e^{\beta h_{ja}} + 1 )} \right\}, \label{eq:BP2}
\end{eqnarray}
where $\partial i \backslash a$ means the nodes in the neighbourhood of $i$ excepting $a$.

We explain how to derive BP equations briefly.
From partition function (\ref{eq:partition}), BP rules are written as
\begin{align}
\nu_{i \to a} (x_i) & = \prod_{b \in \partial i \backslash a} \hat{v}_{b \to i}(x_i) e^{-\mu \beta c_i x_i}, \\
\hat{\nu}_{a \to i} (x_i) & = \sum_{x_j;j \in \partial a \backslash i} \sum_{y_a} \theta \left( \sum_{k \in \partial a}
x_k - y_a \right) e^{\beta w_a y_a} \prod_{j \in \partial a \backslash i} \nu_{j \to a} ( x_j ),
\end{align}
where $\nu_{i \to a}(x_i), \hat{\nu}_{a \to i}(x_i)$ are beliefs in the original equations.
Let us redefine the beliefs by the exponential form,
\begin{align}
\nu_{i \to a} (x_i) \propto e^{\beta h_{ia} x_i}, \hat{\nu}_{a \to i} (x_i) \propto e^{\beta \hat{h}_{ai} x_i}.
\end{align}
By computing the ratio of beliefs between $x_i=0,1$,
\begin{align}
\frac{\nu_{i \to a} (0) }{\nu_{i \to a} (1)} & = e^{-\beta h_{ia}} = e^{ - \beta \sum_{b \in \partial i \backslash a} \hat{h}_{bi} + \mu \beta c_i},
\end{align}
which gives Eq.(\ref{eq:BP1}). Similarly, from the ratio of $\hat{\nu}_{a \to i} (x_i)$,
\begin{align}
\frac{\hat{\nu}_{i \to a} (0) }{\hat{\nu}_{i \to a} (1)} = e^{- \beta \hat{h}_{ia}} 
& = \frac{(1+e^{\beta w_a}) \prod_{j \in \partial a \backslash i} (e^{\beta h_{ja}} + 1) - e^{\beta w_a}}
{(1+e^{\beta w_a }) \prod_{j \in \partial a \backslash i} (e^{\beta h_{ja}} + 1)}.
\end{align}
This yields Eq.(\ref{eq:BP2}) after taking logarithm.

After having beliefs, we calculate the local fields from beliefs,
\begin{eqnarray}
h_{i} &=& - \mu c_i + \sum_{b \in \partial i} \hat{h}_{bi}, \label{eq:BP3} \\
\eta_{a} &=& -\frac{1}{\beta} \ln \left\{ 1 - \frac{1}{1+e^{-\beta w_a}} \frac{1}{\prod_{j \in \partial a}
(e^{\beta h_{ja}} + 1 )} \right\}, \label{eq:BP4}
\end{eqnarray}
and the probability (\ref{eq:prob}) is calculated by these fields.
Accordingly, we can select nodes in $\cal X$ from the values of these fields.

The BP-based algorithm for MC is summarized in algorithm \ref{alg2}.
In this algorithm, the node of the largest $h_i/c_i$ is selected from the remaining ones
like g-greedy algorithm.
Note that the constraint $\sum_i c_i x_i \le K$ is not directly considered in BP formulation. Therefore
we introduce this constraint by combining BP with algorithm \ref{alg1}.\cite{TMH}

\begin{algorithm}                      
\caption{BP-based MC algorithm}         
\label{alg2}                    
\begin{algorithmic}                  
\STATE{initialize beliefs $h_{ia}, \hat{h}_{ai} \forall (a,i) \in {\cal E}$}
\STATE{initialize two sets, $\hat{\cal X} = \{1,\ldots N\},\hat{\cal X}_{\rm cov}= \phi$}
\REPEAT
\STATE{update $h_{ia}\ \forall (a,i) \in {\cal E}$ by Eq.(\ref{eq:BP1})}
\STATE{update $\hat{h}_{ai}\ \forall (a,i) \in {\cal E}$ by Eq.(\ref{eq:BP2})}
\UNTIL{it reaches maximum number of BP iteration}
\STATE{calculate $h_{i}\ \forall i$ by Eq.(\ref{eq:BP3})}
\WHILE{$\hat{\cal X} \ne \phi$}
\STATE{$k = {\rm argmax}_{i \in \hat{\cal X}} ( h_i / c_i )$}
\IF{$c_k + \sum_{i \in \hat{\cal X}_{\rm cov} } c_i \le K$}
\STATE{add $k$ to $\hat{\cal X}_{\rm cov}$}
\ENDIF
\STATE{delete $k$ from $\hat{\cal X}$}
\ENDWHILE
\STATE{output $\hat{\cal X}_{\rm cov}$ }
\STATE{output $\sum_{a \in \partial \hat{\cal X}_{\rm cov}} w_a$ }
\end{algorithmic}
\end{algorithm}


We apply our algorithm to MC on weighted biregular random graph
for verification of MC performance.
In this experiment, we use (9,3)-biregular random graph:
we randomly assign 9 edges for the nodes in $\cal X$, and
3 edges in $\cal Y$. We set the number of nodes as
$N=100$ and $M=300$. For weight $c_i$ and $w_a$, we assign random integer 
number from $1$ to $10$ uniformly.
For BP, parameters are fixed as $\beta=3$ and $K=100$,
and $\mu$ is varied. BP iteration in Algorithm \ref{alg2} is performed 150 times.
We checked the convergence of beliefs after 150 iterations.

\begin{figure}
\begin{picture}(0,135)
\put(20,-8){\includegraphics[width=0.40\textwidth]{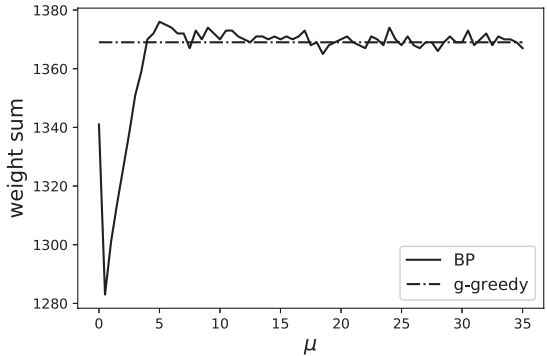}}
\end{picture}
\caption{The result of MC on biregular random graph.}
\label{f2}
\end{figure}

The MC results by g-greedy algorithm and BP algorithm
are shown in Fig.\ref{f2}, where the results are averaged over 1000 random graphs.
As indicated in the case of unweighted graph, the maximal weight sum exceeds the result of
g-greedy algorithm near $\mu=0$. In the present case, the peak of the weight sum is $\mu \simeq 5$. 

Next we apply our algorithm to TS problem by performing MC for weight of word.
In our experiment, we use task2 in DUC2004 dataset\cite{DUC4}.
The dataset consists of 50 clusters of news articles 
from Associated Press and The New York Times, where each cluster has 10 documents. Our task is
to make summarization text from multiple documents in each cluster.  
As references, summarization texts written by human are attached to each cluster.

The weight of word is assigned by Term Frequency-Inverse Document Frequency (TF-IDF)\cite{TFIDF}.
TF-IDF is the product of two factors, TF and IDF:
 a word has high TF-IDF when it appears very frequently (TF) and in very specific sentences in the documents (IDF).
We assign 1.5 times TF-IDF weight to the words in the first sentence of the document, because the first sentence has significant meaning in the document.   
For computing weights by TF-IDF, we also use DUC2003 dataset in addition to DUC2004.
As preprocessing of documents, we use stemming, deletion of exclamation mark and parenthesis, and
conversion of letters to lowercase.

TS performance is evaluated by comparing summarization with the attached reference. Quantitatively, the performance is measured
by ROUGE\cite{ROUGE}, more precisely ROUGE-1. ROUGE-1 is computed by
\begin{equation}
{\rm ROUGE}-1 = \frac{|{\rm words\ in\ summarization} \cap {\rm words\ in\ reference} |} {| {\rm words\ in\ reference}|}.
\end{equation}
Namely, it measures how many words appear commonly both in summarization and reference.
In this experiment, we take average of ROUGE-1 over 50 clusters and 4 attached references for each cluster.
For evaluation of ROUGE-1, we use the tool SumEval\cite{SumEval}.

\begin{figure}[t]
\begin{picture}(0,280)
\put(28,128){\includegraphics[width=0.385\textwidth]{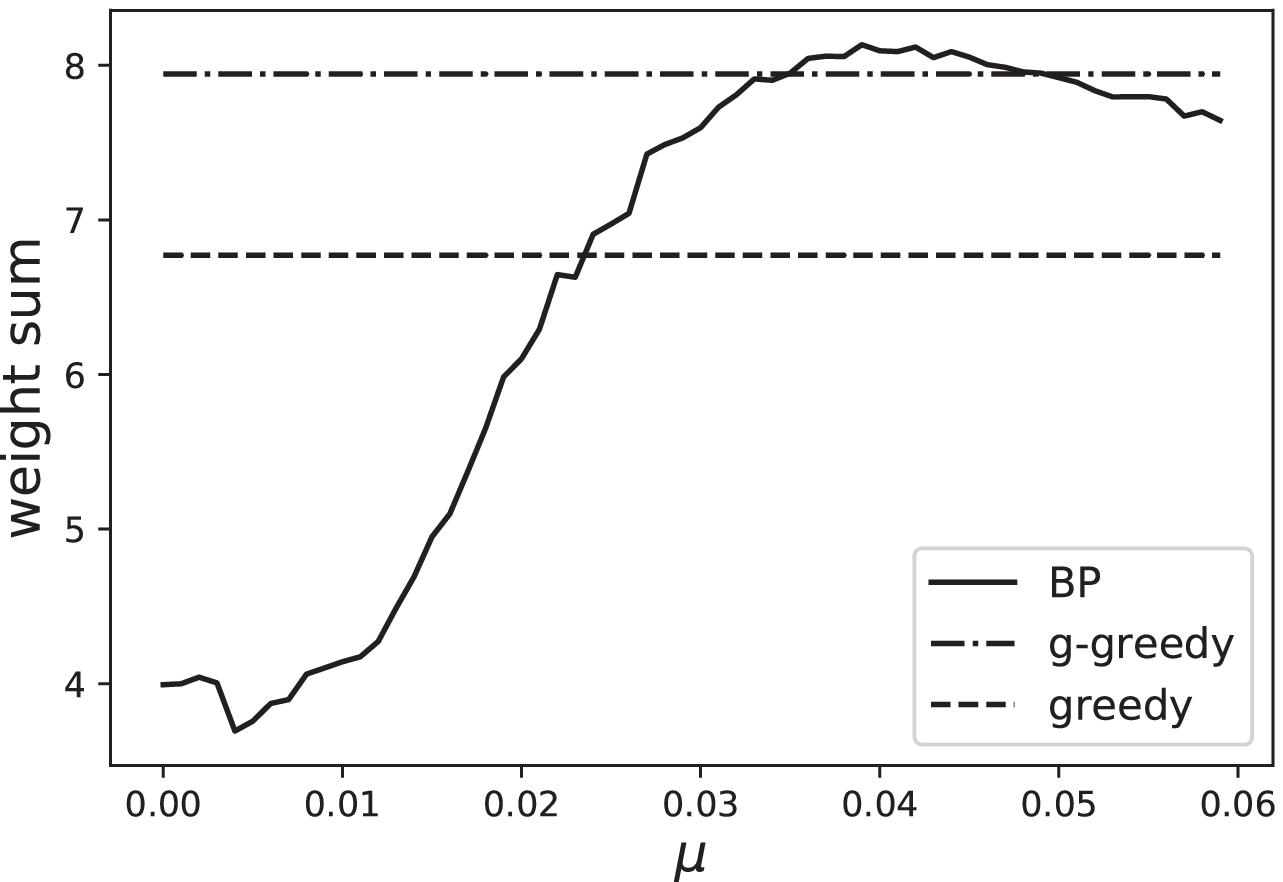}}
\put(20,-10){\includegraphics[width=0.40\textwidth]{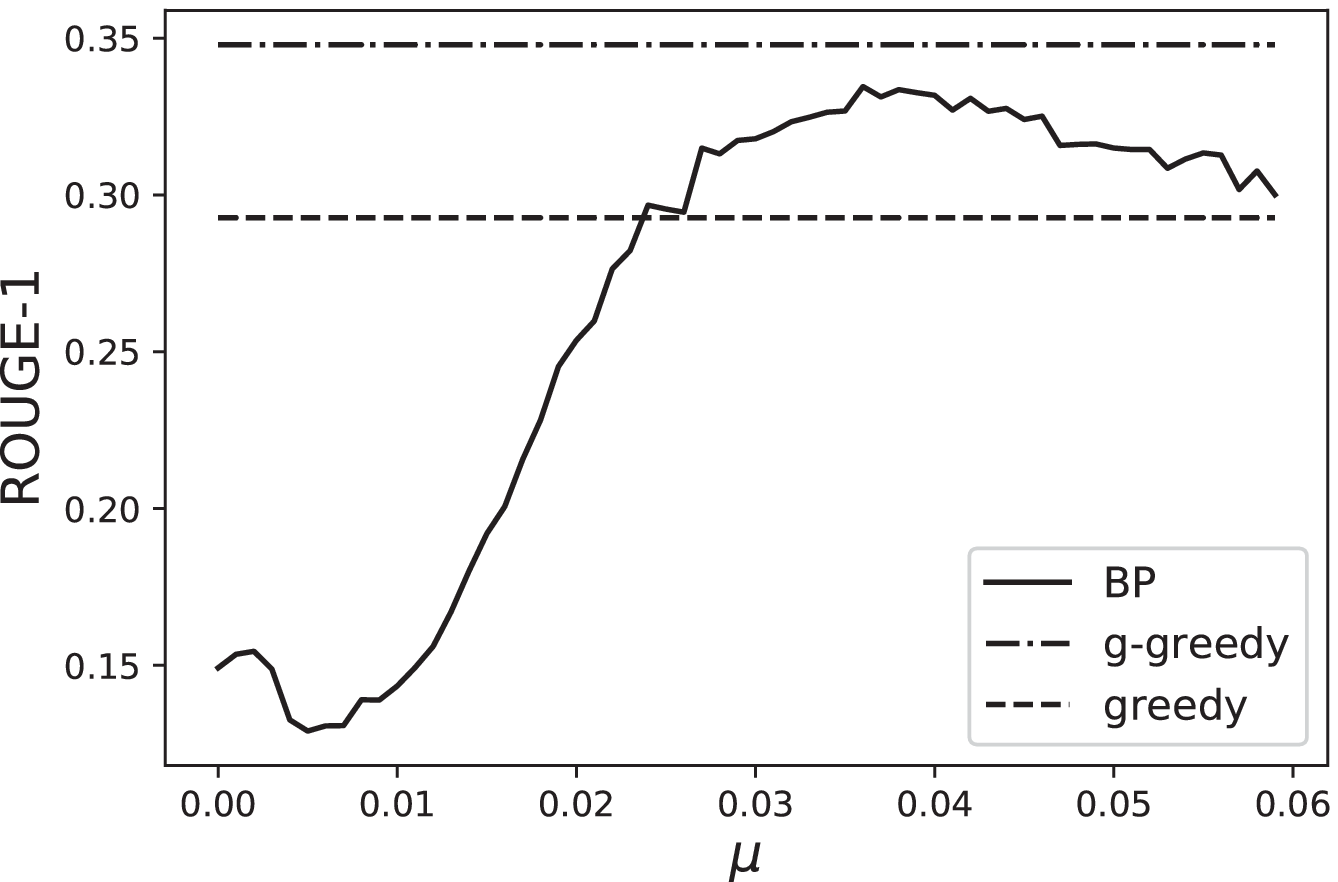}}
\end{picture}
\caption{The result of TS for DUC2004 dataset with removal of stop words.
Top: weight sum of covered nodes. Bottom: ROUGE-1.}
\label{f3}
\end{figure}

\begin{figure}[t]
\begin{picture}(0,280)
\put(28,128){\includegraphics[width=0.385\textwidth]{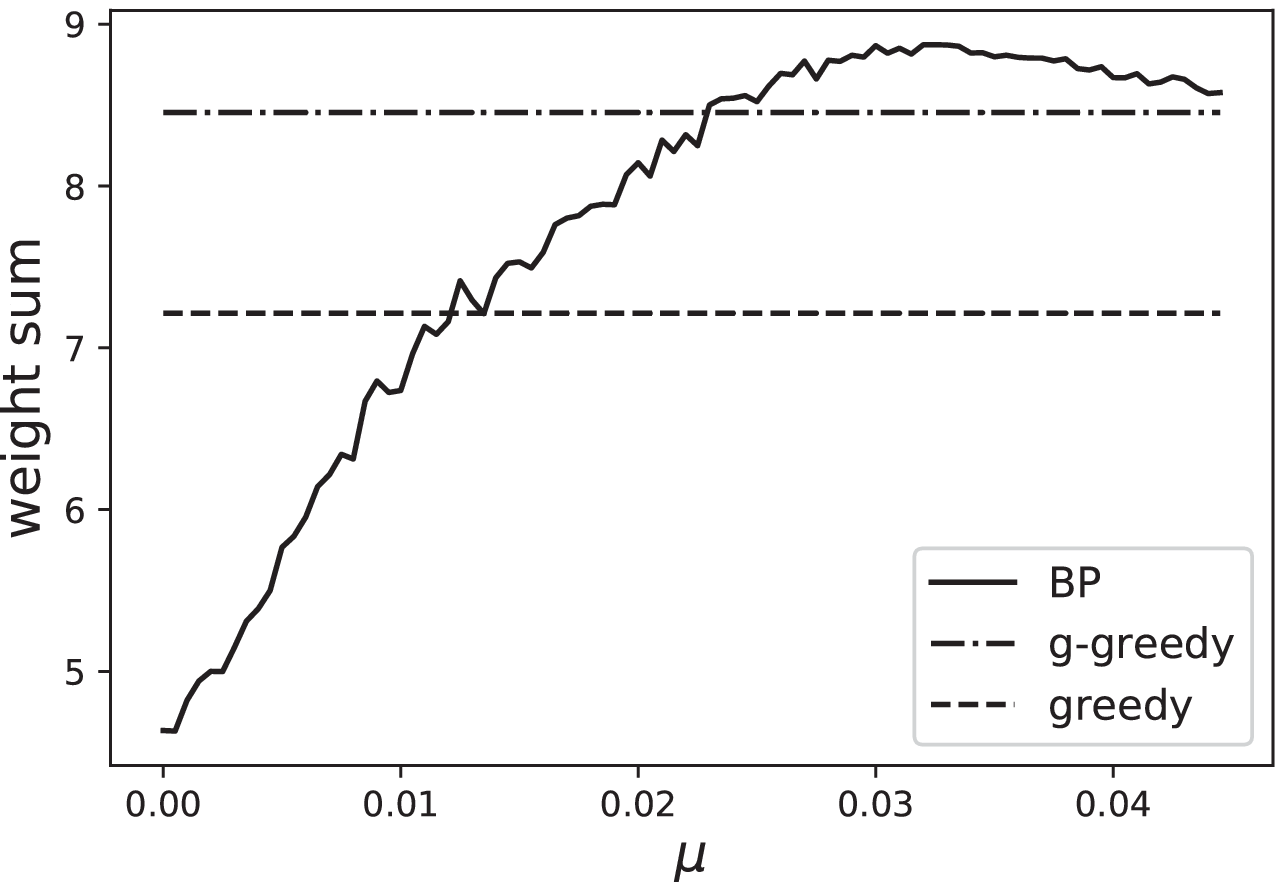}}
\put(20,-10){\includegraphics[width=0.40\textwidth]{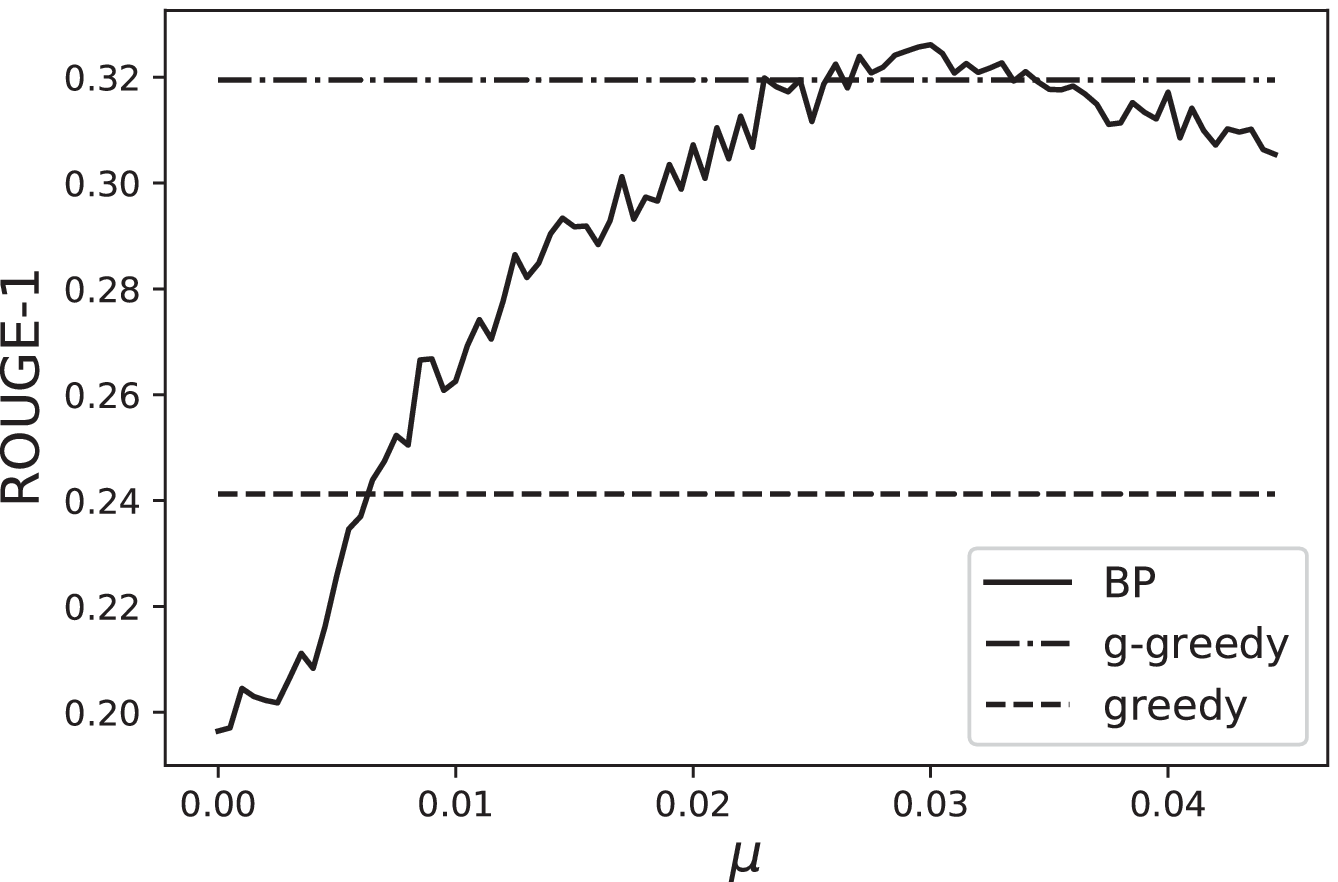}}
\end{picture}
\caption{The result of TS for DUC2004 dataset without removal of stop words.
Top: weight sum of covered nodes. Bottom: ROUGE-1.}
\label{f4}
\end{figure}

We show two results. In our experiment $K=100$.
BP iteration is performed 150 times, and we checked the convergence of beliefs after 150 iterations.
The first result is depicted in Fig.\ref{f3}, where $\beta=45$.
In this result, we remove stop words from documents by Natural Language Toolkit\cite{NLTK}:
stop words are prepositions and articles such as "a" "the".
There is a peak of weight sum at $\mu \simeq 0.04$, and
ROUGE-1 shows the peak at almost the same $\mu$. However, our algorithm does not outperform g-greedy in terms of ROUGE-1.
The second result is in Fig.\ref{f4}, where $\beta=80$ and stop words are not removed. 
In this result, the maximal ROUGE-1 exceeds the value of g-greedy
around the peak $\mu \simeq 0.03$. 
We also change the value of $\beta$ within the range $10 \le \beta \le 100$, and the best $\beta$ is used in 
Figs.\ref{f3},\ref{f4}. In this dataset, typical value of the weight is small, $w_a \simeq 10^{-2}$. Then the appropriate value of $\beta$
should be large for satisfying $ \beta \sum_{a} w_a y_a \sim 1$ in Eq.(\ref{eq:partition}).

As a consequence, maximal ROUGE-1 is larger than the one of g-greedy when stop words are not removed, while the result is worse by removal
of stop words.
We expect the reason is that the inclusion of stop words affects the weight of word.
By including stop words, the precision of TF-IDF weight might be statistically improved
by larger number of words.
However, we should also keep in mind that stop words are often excluded in natural language processing.


To summarize, we generalized BP-based MC algorithm for weighted graph. Then we applied our algorithm to MC on weighted random graph,
and had better performance than g-greedy.
We also applied it to TS, whose result indicates 
that the advantage over g-greedy depends on the weight of words. 
As future work, we should investigate in what cases it exhibits better performance than g-greedy in further detail.

\vspace{2mm}
\begin{acknowledgment}

We are thankful to Satoshi Takabe for discussion and helpful comments. This work is supported by KAKENHI Nos. 18K11175, 19K12178.

\end{acknowledgment}



\begin{thebibliography}{9}
\bibitem{McDoland} R. McDonald, Proc. of the 29th European Conference on Information Retrieval, 557-564 (2007).
\bibitem{Filatova} E. Filatova and V. Hatzivassiloglou, Proc. of the 20th International Conference on Computational Linguistics, 397-403 (2004).
\bibitem{TO} H. Takamura and M. Okumura, Proc. of the 12th Conference of the European Chapter of the ACL, 781-789 (2009). 
\bibitem{TMH} S. Takabe, T. Maehara, and K. Hukushima, Phys Rev. E {\bf 97}, 022138 (2018).
\bibitem{KMN} S. Khuller, A. Moss, and J. S. Naor, Information Processing Letters {\bf 70}, 39-45 (1999).
\bibitem{DUC4} Document Understanding Conference, NAACL-HLT Workshop on Text Summarization (2004).
\bibitem{TFIDF} S. Gerard, {\it Automatic Text Processing} (Addison-Wesley, Reading, 1989).
\bibitem{ROUGE} C.-Y. Lin and E. Hovy, Proc. of the 4th meeting of the NAACL-HLT, 150-157 (2003).
\bibitem{SumEval} https://github.com/chakki-works/sumeval
\bibitem{NLTK} http://www.nltk.org/

\end{thebibliography}
\end{document}